\documentclass[conference,letterpaper]{IEEEtran}
\IEEEoverridecommandlockouts
\usepackage{cite}
\usepackage{amsmath,amssymb,amsfonts}
\usepackage{algorithmic}
\usepackage{graphicx}
\usepackage{textcomp}
\usepackage{xcolor}
\def\BibTeX{{\rm B\kern-.05em{\sc i\kern-.025em b}\kern-.08em
    T\kern-.1667em\lower.7ex\hbox{E}\kern-.125emX}}

\ifCLASSOPTIONcompsoc
  \usepackage[caption=false,font=normalsize,labelfont=sf,textfont=sf]{subfig}
\else
  \usepackage[caption=false,font=footnotesize]{subfig}
\fi

\begin{document}

\title{Hybrid Mamba-Attention Neural Architecture for Channel Estimation}

\author{
\IEEEauthorblockN{Dianxin Luan\IEEEauthorrefmark{1}, Chengsi Liang\IEEEauthorrefmark{2}, Jie Huang\IEEEauthorrefmark{3}\IEEEauthorrefmark{4}, Zheng Lin\IEEEauthorrefmark{5}, Kaitao Meng\IEEEauthorrefmark{6}, Cheng\mbox{-}Xiang Wang\IEEEauthorrefmark{3}\IEEEauthorrefmark{4}, John Thompson\IEEEauthorrefmark{1} \\ and Ozgur Akan \IEEEauthorrefmark{7}}
\IEEEauthorblockA{\IEEEauthorrefmark{1} Institute for Imaging, Data and Communications, School of Engineering, University of Edinburgh, Edinburgh, UK}
\IEEEauthorblockA{\IEEEauthorrefmark{2} James Watt School of Engineering, University of Glasgow, Glasgow, UK}
\IEEEauthorblockA{\IEEEauthorrefmark{3} National Mobile Communications Research Laboratory, Southeast University, Nanjing 211189, China}
\IEEEauthorblockA{\IEEEauthorrefmark{4} Purple Mountain Laboratories, Nanjing 211111, China}
\IEEEauthorblockA{\IEEEauthorrefmark{5} Department of Electrical and Electronic Engineering, The University of Hong Kong, Hong Kong, China}
\IEEEauthorblockA{\IEEEauthorrefmark{6} Department of Electrical and Electronic Engineering, University of Manchester, Manchester, UK}
\IEEEauthorblockA{\IEEEauthorrefmark{7} Internet of Everything Group, Department of Engineering, University of Cambridge, UK} 
\IEEEauthorblockA{\textit{Emails:} \{dianxin.luan, john.thompson\}@ed.ac.uk, 2357875l@student.gla.ac.uk \\
\{j.huang, chxwang\}@seu.edu.cn \\ linzheng@eee.hku.hk, kaitao.meng@manchester.ac.uk, oba21@cam.ac.uk} 
}

\maketitle

\begin{abstract}
This paper proposes a hybrid Mamba-attention neural architecture to achieve improved channel estimation for orthogonal frequency-division multiplexing (OFDM) waveforms, particularly for configurations with a large number of subcarriers. By integrating a customized Mamba module, the proposed framework handles large-scale subcarrier channel estimation efficiently while capturing long-distance dependencies among these subcarriers effectively. Unlike the conventional Mamba structure, this paper implements a bidirectional selective scan to enable information propagation from both directions, because channel gains at different subcarriers are inherently non-causal. In addition, by integrating Mamba to reduce the reliance on quadratic-complexity self-attention, the proposed solution achieves lower space complexity than fully transformer architectures. Simulation results based on the 3GPP TS 36.101 channel demonstrate that compared to other baseline neural networks, the proposed method achieves superior channel estimation performance with fewer tunable parameters and exhibits good generalization across previously unseen channels. 
\end{abstract}

\section{Introduction}
For sixth generation (6G) communication systems, orthogonal frequency-division multiplexing (OFDM) is expected to continue to serve as baseband modulator \cite{wang2023road, wu2024intelligent, lu2024tutorial, wang2020artificial}. Therefore, acquiring accurate channel state information (CSI) is essential as it ensures effective compensation for channel impairments, supports high data rate and reliable connectivity. Conventional approaches, such as least squares (LS) and minimum mean squared error (MMSE) \cite{van1995channel, edfors1998ofdm}, cannot meet current demands. The LS estimator has poor performance, whereas the MMSE method, despite its theoretical optimality, relies on actual channel information that is unavailable in the real world. In addition, precise interpolation for doubly-selective channels is usually challenging because statistical models are inaccurate and difficult to handle \cite{nissel2018doubly}, especially for channel estimates at the data OFDM symbols. These drawbacks highlight the necessity of alternative solutions. Therefore, artificial intelligence (AI) solutions are attracting significant attention due to their ability to achieve superior performance under complex propagation conditions \cite{ye2017power, neumann2018learning, nerini2022machine, fesl2024diffusion, guo2024parallel, 11090043, 11359105, yuan2020bayesian}. 

Unlike conventional methods that emphasize closed-form expressions, neural networks aim for locally optimal solutions, such as ChannelNet \cite{soltani2019deep} and ReEsNet \cite{li2019deep}. However, neural networks suffer from significant degradation on new channels. To fine-tune neural networks online, the paper \cite{kong2025representation} proposes a continuous learning-based channel estimation scheme that preserves historical knowledge and achieves an ongoing process of self-improvement. Moreover, the paper \cite{wang2025fedpda} proposes federated collaborative learning with pruned-data aggregation to reduce the frequency of online adaptation. To achieve robust generalization, our works \cite{luan2023achieving, luan2025achieving} propose design criteria to generate synthetic training datasets that ensure trained channel estimation neural networks perform robustly on previously unseen channels. With the attention-based architecture released in \cite{vaswani2017attention}, the paper \cite{liu2023vit, liu2024pd} proposes an attention-based channel estimator with a pilot switch network that adapts configurations to channel conditions, balancing accuracy and overhead. An efficient parallel Transformer is proposed in \cite{guo2024parallel} to achieve efficient channel estimation by considering estimation time for reconfigurable intelligent surface (RIS)-aided wireless communication. Our prior studies \cite{luan2022attention, luan2023channelformer} also propose an encoder–decoder architecture that leverages the transformer encoder to achieve input sparsity according to their relative importance. However, although transformer-based neural networks achieve improved performance, their computational and spatial complexities increase rapidly with the number of OFDM subcarriers due to the larger input sequence length. This will result in high computational consumption, challenging battery-powered handset devices. 

In this paper, we propose a hybrid Mamba-attention architecture for channel estimation in OFDM systems with a large number of subcarriers. Unlike transformer-based approaches that suffer from quadratic complexity and lack inductive bias for structured channel correlations, the proposed framework leverages a state-space model to efficiently capture long-range dependencies. To address the inherent non-causal correlations across subcarriers, a bidirectional selective state scan is introduced to enable information propagation from both directions. Simulation results based on the 3GPP TS 36.101 channel demonstrate that the proposed architecture achieves improved performance with fewer tunable parameters compared to existing neural network-based channel estimators. 
\section{System architecture}
\label{System architecture and propagation channels}
\subsection{Baseband and channel settings}
This paper considers single antenna transmission for an OFDM cellular system with a frequency spacing of $f_{space}$ at carrier frequency $f_{r}$. The randomly generated source signal $\mathbf{s}$ is processed by a Quadrature Phase Shift Keying (QPSK) modulator. The QPSK modulated symbols are assigned to the data subcarriers in the slot. Each slot consists of $N_f$ subcarriers and $N_s$ OFDM symbols. By following the 5G NR specifications in 3GPP TS 38.211, the employed demodulation reference signal (DM-RS) pattern is a single-symbol DM-RS with three additional positions as the default pilot pattern. The 3\textsuperscript{rd}, 6\textsuperscript{th}, 9\textsuperscript{th} and 12\textsuperscript{th} OFDM symbols are reserved for pilots ($N_{pilot} = 4$). For each pilot OFDM symbol, the second subcarriers of each indices of $L_s = 4$ subcarriers is reserved for pilot subcarriers. The pilot signals are assigned by a fixed value, and the vacant pilot subcarriers are set to 0. The inverse fast Fourier transform converts the slot signal into the time domain. Normally the fast Fourier transform (FFT) and inverse fast Fourier transform (IFFT) operators use scaling factors of 1 and $1/{N_f}$, but those are changed to $1/\sqrt{N_f}$ to avoid changing the power of their outputs. Then the Cyclic-Prefix (CP, of length $L_{\mathrm{CP}}$ samples) is added to the front of each symbol to resist the multipath fadings. The channel is assumed to be a multipath channel with $M$ paths which has an impulse response of 
\begin{equation}
h(\tau, t) = \sum_{m=0}^{M-1} a_m\left(t\right)\delta(\tau-\tau_mT_s) 
\label{impulse response}
\end{equation}
where $a_m\left(t\right) = a_me^{-j2\pi f_{D,m}t+\phi_{D,m}}$ is the path gain of the $m$th path, $\tau_m$ is the corresponding delay normalized by the sampling period $T_{s}$, $f_{D,m}$ is the corresponding Doppler shifts and $\phi_{D,m}$ is the corresponding initial phase. Therefore, the $n$th sampled time-domain $\mathbf{h}(n; t)$ is 
\begin{equation}
    \mathbf{h}(n; t) = \sum_{m=0}^{M-1} a_m\left(t\right)e^{-j\pi\frac{n+(N_f-1)\tau_m}{N_f}}\frac{\sin(\pi\tau_m)}{\sin(\frac{\pi}{N_f}(\tau_m-n))}. 
\label{channel tap}
\end{equation}

For the propagation channel model, this paper uses the Extended Typical Urban (ETU) channel from 3GPP TS 36.101 document, which represents a high delay spread environment. Then the sampled time-domain received signal $\mathbf{y}$ is given by 
\begin{equation}
\mathbf{y} = \mathbf{h}\otimes \mathbf{x} + \mathbf{n} 
\label{h}
\end{equation}
where $\otimes$ denotes the convolution operation, $\mathbf{n}$ is assumed to be additive white Gaussian noise (AWGN) and $\mathbf{x}$ is the transmitted time-domain OFDM signal. Each slot is assigned a new channel realization. After removing the CP, the receiver converts the time-domain data to the frequency-domain using the FFT operation. By assuming that $\forall{\tau_{m}} \leq L_{CP}$, the channel gain at the $k$th subcarrier is 
\begin{equation}
\mathbf{H}\left(k, l\right) = \frac{1}{\sqrt{N_f}}\sum_{m=0}^{M-1} a_me^{-j2\pi \left(f_{D,m}T_ol+\phi_{D,m} +\frac{k\tau_m}{N_f}\right)}. 
\label{Phase}
\end{equation}
\normalsize
for $k = [0, N_f-1]$ and $T_o$ is one complete OFDM symbol period including the CP. The received signal at $k$th subcarrier and $l$th OFDM symbol, $\mathbf{Y}(k, l)$, is of the form: 
\begin{equation}
\mathbf{Y}(k, l) = \mathbf{H}(k, l)\mathbf{X}(k, l) + \mathbf{N}(k, l) 
\end{equation}
where $\mathbf{N} \sim \mathcal{CN}(\mathbf{0}, \ \sigma_{N}^{2}\mathbf{I}_{N_f})$, and complete $\mathbf{H} \in \mathbb{C}^{{N_f}\times N_{s}}$ is the channel matrix to be estimated and $\sigma_X^2 = E\{(N_fN_s)^{-1}\left\Arrowvert\mathbf{X}\right\Arrowvert_{F}^{2}\}$. The received pilot signal is then extracted to provide a reference for the channel matrix of each packet. The recovered signal will be filtered to remove the channel effects, and then processed in a QPSK demodulator to obtain the received bit-level data estimates $\mathbf{\hat{s}}$. 
\subsection{Conventional methods and baseline neural networks}
The LS estimate at the $k$th subcarrier is given by 
\begin{equation}
    \mathbf{\hat{H}_{ls}}(k) = Y_{k}X_{k}^{-1}, 
\end{equation}
where $Y_{k}$ and $X_{k}$ denote the received and transmitted pilot signals for the $k$th subcarrier, respectively. This paper applies linear interpolation on $\mathbf{\hat{H}_{ls}^{pilot}} \in \mathbb{C}^{\frac{N_f}{L_{s}}\times N_{pilot}}$ (LS estimate at the pilot positions) to predict the complete channel matrix. By utilizing exactly actual channel information for improved estimate performance, the linear MMSE estimate for one OFDM symbol $\mathbf{\hat{H}_{MMSE}} \in \mathbb{C}^{N_f}$ is computed by 
\begin{equation}
    \mathbf{\hat{H}_{MMSE}} = \mathbf{R_{H_cH_p}}\left(\mathbf{R_{H_pH_p}} + \left(\frac{\sigma_N^2}{\sigma_X^2}\right) \mathbf{I}\right)^{-1}\mathbf{\hat{H}_{ls}} 
\label{MMSE}
\end{equation}
where $\mathbf{H_{c}} \in \mathbb{C}^{N_f}$ is the actual channel matrix for each OFDM symbol and $\mathbf{H_p} \in \mathbb{C}^{\frac{N_f}{L_{s}}}$ is the actual channel matrix at the corresponding pilot subcarriers, which are computed by 
\begin{equation}
    \mathbf{R_{H_{c}H_p}} = \mathbb{E}\left\{\mathbf{H_{c}}\mathbf{H_{p}^{H}}\right\}, \mathbf{R_{H_pH_p}} = \mathbb{E}\left\{\mathbf{H_{p}H_{p}^{H}}\right\}. 
\end{equation}

These matrices are substituted into (\ref{MMSE}) to calculate the $\mathbf{\hat{H}_{MMSE}^{pilot}} \in \mathbb{C}^{N_f \times N_{pilot}}$. $\mathbf{\hat{H}_{MMSE}^{pilot}}$ is then interpolated linearly to obtain the estimate for the complete slot. To evaluate performance, InterpolateNet \cite{luan2021low}, HA02 \cite{luan2022attention} and Channelformer \cite{luan2023channelformer} are implemented as baseline neural networks. InterpolateNet is a low-complexity convolutional neural network with skip connections that only has 9,442 tunable parameters. Both HA02 and Channelformer exploit sub-channel correlations among the input features for improvement. 
\section{MambaNet: a hybrid Mamba-attention architecture}
\label{Mamba-assisted framework}
We propose a hybrid Mamba-attention architecture (called MambaNet), which implements the multi-head attention mechanism to capture the correlation among the LS input, and then uses a customized Mamba structure to efficiently propagate and refine the channel correlation information over a very large number of subcarriers (large $N_f$), leading to both improved performance and reduced complexity. 

For the input of MambaNet, the $\mathbf{\hat{H}_{LS}^{pilot}} \in \mathbb{C}^{\left(\frac{N_f}{L_s}\right) \times N_{pilot}}$ is concatenated to be one column vector $\in \mathbb{C}^{\frac{N_f}{L_s}N_{pilot}}$. The real and imaginary parts of this vector are split into two channels as the second dimension with a size of $\mathbb{R}^{\left(\frac{N_{pilot}N_f}{L_s}\right) \times 2}$ to form the channel tokenization. The MambaNet is trained by the complete channel matrix $\in \mathbb{R}^{{N_f} \times N_{s} \times 2}$ where the second dimension stores the real and imaginary part of the complex channel matrix for the slot. 
%
%
\subsection{Attention-assisted Mamba module}
To ensure neural networks effectively capture the correlation among different subcarriers and OFDM symbols, it consists of two components which are self-attention module and Mamba module. The first fully-connected layer of the multi-head attention module resizes the LS input $\in \mathbb{R}^{\left(\frac{N_{pilot}N_f}{L_s}\right) \times 2}$ to $\in \mathbb{R}^{\left(\frac{3N_{pilot}N_f}{L_s}\right) \times 2}$ by a linear projection 
\begin{equation}
    \mathbf{y} = \mathbf{Wx} + \mathbf{b} 
\end{equation}
where $\mathbf{W} \in \mathbb{R}^{\left(\frac{3N_{pilot}N_f}{L_s}\right) \times \left(\frac{N_{pilot}N_f}{L_s}\right)}$, $\mathbf{b} \in \mathbb{R}^{\left(\frac{3N_{pilot}N_f}{L_s}\right) \times 1}$ are tunable parameters. Multi-head attention equally splits that to obtain $\mathbf{K}$, $\mathbf{Q}$ and $\mathbf{V}$ $\in \mathbb{R}^{\left(\frac{N_f}{L_s}\right) \times 2 \times N_{head}}$ for $N_{head} = N_{pilot}$ heads. For each head, the scaled dot-product attention $\mathbf{Y_{\mathrm{Attention}}} \in \mathbb{R}^{\left(\frac{N_f}{L_s}\right) \times 2 \times N_{heads}}$ \cite{vaswani2017attention} is computed by 
\begin{equation}
    \mathbf{Y_{\mathrm{Attention}}} = \mathrm{softmax}\left(\frac{\mathbf{QK}^{T}}{\sqrt{\frac{N_f}{L_s}}}\right)\mathbf{V}. 
\end{equation}
After concatenating the scaled dot-product attention for each head in the first dimension, the concatenated result $\in \mathbb{R}^{\left(\frac{{N_f}{N_{heads}}}{L_s}\right) \times 2}$ is then processed by a fully-connected layer to generate the output $\in \mathbb{R}^{\left(\frac{{N_f}{N_{heads}}}{L_s}\right) \times 2}$. The layer normalization module then transformed the superimposed result $x\in \mathbb{R}^{\left(\frac{{N_f}{N_{pilot}}}{L_s}\right) \times 2}$ of this output and the neural network input, to obtain the output of the multi-head attention $ y\in \mathbb{R}^{\left(\frac{{N_f}{N_{pilot}}}{L_s}\right) \times 2}$, by 
\begin{equation}
    \mathbf{y} = \mathbf{w}\left(\frac{\mathbf{x} - \mu}{\sqrt{\sigma_x^{2} + \varepsilon}}\right) + \mathbf{b} 
\end{equation}
where $\mu$ is the mean value of the input $\mathbf{x}$ and $\sigma_x^{2}$ is the corresponding variance, $\varepsilon = 10^{-5}$ to avoid zero-division and $\mathbf{w}, \ \mathbf{b} \in \frac{N_{pilot}N_f}{L_s} \times 1$ are the tunable parameters. 

We adopt a modified Mamba module operating on the multi-head attention output $\in \mathbb{R}^{\left(\frac{{N_f}{N_{pilot}}}{L_s}\right) \times 2}$ to further propagate channel correlations across a large number of subcarriers with low complexity. The Mamba module is based on the selective space state model (SSM) consists of two main components: gate coefficients generation for dynamic control and selective state scan for state updates. Let the input of Mamba module be denoted as $\mathbf{X^{\text{mamba}}} \in \mathbb{R}^{\left(\frac{N_{pilot}N_f}{L_s}\right)\times 2}$, where the first dimension corresponds to the effective sequence length of the attention signals and the second dimension denotes the channel dimension. A fully connected layer is first applied to increase the channel dimension of $\mathbf{X^{\text{mamba}}}$ from $\mathbb{R}^{\left(\frac{N_{pilot}N_f}{L_s}\right)\times 2}$ to $\mathbb{R}^{\left(\frac{N_{pilot}N_f}{L_s}\right)\times 2C_{spread}}$ where $C_{spread}$ = 24, and then split this output to two sub-matrix $\mathbf{U_{main}} \in \mathbb{R}^{\left(\frac{N_{pilot}N_f}{L_s}\right)\times C_{spread}}$ and $\mathbf{U_{gate}} \in \mathbb{R}^{\left(\frac{N_{pilot}N_f}{L_s}\right)\times C_{spread}}$. 
\subsubsection{Gate coefficients generation}
To calculate the gate coefficients, depthwise convolution is applied to $\mathbf{U_{main}}$ to obtain $\mathbf{X_c} \in \mathbb{R}^{\left(\frac{N_{pilot}N_f}{L_s}\right)\times C_{spread}}$ for no cross-channel mixing. The LS estimate is inherently non-causal, resulting in subsequent in-process features remaining non-causal. Then the silu layer is applied to $\mathbf{X_c} \in \mathbb{R}^{\left(\frac{N_{pilot}N_f}{L_s}\right)\times C_{spread}}$, defined as 
\begin{equation}
    \mathbf{X_{dc}} = \mathrm{silu}\left(\mathbf{X_c}\right) = \frac{\mathbf{X_c}}{1 + e^{-\mathbf{X_c}}}. 
\end{equation}

The $\mathbf{X_{dc}} \in \mathbb{R}^{\left(\frac{N_{pilot}N_f}{L_s}\right)\times C_{spread}}$ is then projected into three branches to generate the gate coefficients $\mathbf{a}, \mathbf{b}, \mathbf{g} \in \mathbb{R}^{\left(\frac{N_{pilot}N_f}{L_s}\right)\times C_{spread}}$ by 
\begin{align}
\mathbf{a} &= \sigma(\mathbf{X_{dc}}\mathbf{W_a} + \mathbf{b_a}), \mathbf{W_a} \in \mathbb{R}^{24\times 24}, \mathbf{b_a} \in \mathbb{R}^{24}, \\ 
\mathbf{b} &= \mathbf{X_{dc}}\mathbf{W_b} + \mathbf{b_b},         \mathbf{W_b} \in \mathbb{R}^{24\times 24}, \mathbf{b_b} \in \mathbb{R}^{24}, \\
\mathbf{g} &= \mathrm{silu}(\mathbf{U_{gate}}\mathbf{W_g} + \mathbf{b_g}), \mathbf{W_g} \in \mathbb{R}^{24\times 24}, \mathbf{b_g} \in \mathbb{R}^{24}, 
\end{align}
where $\sigma(\cdot) = \frac{1}{1+e^{-x}}$ is the sigmoid activation. 
\subsubsection{Bidirectional selective state space scan}
The coefficients $\mathbf{a}, \mathbf{b}$ are used to calculate the scan recurrence for state updates selectively. Because the channel gain elements are non-causal, a backward scan is deployed to effectively capture the non-causal correlation rather than only performing a forward scan. For each step $t=\left[1,\frac{N_{pilot}N_f}{L_s}\right]$, the hidden state $\mathbf{h_t} \in \mathbb{R}^{1 \times C_{spread}}$ is given by 
\begin{align}
\textbf{Forward:}\quad
\mathbf{h^{\mathrm{f}}_{t}} &= \mathbf{a_t} \circ \mathbf{h^{\mathrm{f}}_{t-1}} + \mathbf{b_t}, \quad \mathbf{h^{\mathrm{f}}_{0}} = \mathbf{0}, \\ 
\textbf{Backward:}\quad
\mathbf{h^{\mathrm{b}}_{t}} &= \mathbf{a_t} \circ \mathbf{h^{\mathrm{b}}_{t+1}} + \mathbf{b_t}, \quad \mathbf{h^{\mathrm{b}}_{\left(\frac{N_{pilot}N_f}{L_s}+1\right)}} = \mathbf{0}, \\ 
\textbf{Overall:}\quad
\mathbf{h_t} &= \mathbf{h^{\mathrm{f}}_{t}} + \mathbf{h^{\mathrm{b}}_{t}}. 
\label{state update}
\end{align}
where $\mathbf{a_t}, \mathbf{b_t}$ are the $t$th row of $\mathbf{a}, \mathbf{b}$ and the operator $\circ$ denotes the Hadamard product. As shown in (16) and (17), the coefficient $\mathbf{a}$ controls the degree to which the adjacent hidden state is preserved. A higher value of $\mathbf{a}$ retains more information of the adjacent hidden states, while a lower value accelerates forgetting. The coefficient $\mathbf{b}$ provides an additive contribution that injects new information from the current input into the state update. To compute this recurrence efficiently, (\ref{state update}) is equivalently expressed in parallel form as 
\begin{align}
    &\mathbf{H^{\text{f}}_t} = \left(\prod_{i=1}^{t} \mathbf{a_i}\right) \sum_{i=1}^{t} \frac{\mathbf{b_i}}{\prod_{j=1}^{i} \mathbf{a_j}}, \\ 
    &\mathbf{H^{\text{b}}_t} = \left(\prod_{t}^{\frac{N_{pilot}N_f}{L_s}} \mathbf{a_i}\right) \sum_{t}^{\frac{N_{pilot}N_f}{L_s}} \frac{\mathbf{b_i}}{\prod_{j=i}^{\frac{N_{pilot}N_f}{L_s}} \mathbf{a_j}}, \\ 
    &\mathbf{H^{\text{mamba}}_t} = \mathbf{H^{\text{f}}_t} + \mathbf{H^{\text{b}}_t}, 
\end{align}
leading to stacked states $\mathbf{H^{\text{mamba}}} \in \mathbb{R}^{\left(\frac{N_{pilot}N_f}{L_s}\right)\times C_{spread}}$ for the entire step $t$. The gate control signal $\mathbf{g}$ modulates the entire $\mathbf{H}$ and determines the extent to which the internal state contributes to the output. This output is then projected to obtain the Mamba output $\mathbf{Y^{\text{mamba}}}\in \mathbb{R}^{\left(\frac{N_{pilot}N_f}{L_s}\right)\times 2}$, which is given by 
\begin{equation}
\mathbf{Y^{\text{mamba}}} = \left(\mathbf{g} \circ \mathbf{H^{\text{mamba}}_t}\right)\mathbf{W_{\text{out}}} + \mathbf{b_{\text{out}}} 
\end{equation}
where $\mathbf{W_{\text{out}}} \in \mathbb{R}^{C_{spread}\times 2}, \mathbf{b_{\text{out}}} \in \mathbb{R}^{2}$ are tunable parameters. Then layer normalization is applied to the superimposed results of the Mamba input $\mathbf{X^{\text{mamba}}} \in \mathbb{R}^{\left(\frac{N_{pilot}N_f}{L_s}\right)\times 2}$ and the Mamba output $\mathbf{Y^{\text{mamba}}} \in \mathbb{R}^{\left(\frac{N_{pilot}N_f}{L_s}\right)\times 2}$ to form the output of attention-assisted Mamba module $\mathbf{Y_{a}} \in \mathbb{R}^{\left(\frac{N_{pilot}N_f}{L_s}\right)\times 2}$. 

For time complexity of this Mamba module, linear projections operate on the channel dimension, while the temporal dependencies are modeled entirely through the selective scan along the sequence length (first dimension). It exhibits linear complexity $\mathcal{O}\left(\frac{N_{pilot}N_f}{L_s}\right)$, which is substantially lower than the $\mathcal{O}\left((\frac{N_{pilot}N_f}{L_s})^2\right)$ complexity of self-attention. Therefore, the Mamba module has the capability to model long-range dependencies efficiently, especially when the $N_f$ is very large or the in-processing features of the previous multi-head attention module have a large sequence length. 
\subsection{Residual architecture with up-sampling modules}
To achieve the precise channel estimate for the complete slot, the in-processing features $\mathbf{Y_{a}} \in \mathbb{R}^{\left(\frac{N_{pilot}N_f}{L_s}\right)\times 2}$ are reshaped to $\mathbb{R}^{\left(\frac{N_f}{L_s}\right) \times N_{pilot} \times 2}$. This enlarges the effective receptive field compared to the previous sequence features, which should improve the performance of residual architecture on super-resolution. The first convolutional layer has 12 filters with a kernel size of {5$\times$5$\times$2} followed by a stack of five residual blocks. Each residual block consists of one convolutional layer with 12 filters, each corresponding to a kernel size of {5$\times$5$\times$12}, followed by one ReLU layer and one convolutional layer with 12 filters and the kernel size of each is {5$\times$5$\times$12}. The layer normalization processes the superimposed result of the input and output of the final residual block to the upsampling module. 

This paper exploits bilinear interpolation as up-sampling function, which resizes the output $\in \mathbb{R}^{\frac{N_f}{L_s} \times N_{pilot} \times 12}$ from the previous residual stacks to a size of $\mathbb{R}^{N_f \times N_s \times 12}$. The last coupled convolutional layer has 2 filters with the kernel size of {96$\times$5} to generate the output $\in \mathbb{R}^{{N_f} \times N_{s} \times 2}$. 
\section{Simulations}
\label{Simulations}
MSE is a key performance metric that evaluates the distance between the actual channel matrix and the estimate, which is defined as 
\begin{equation}
    \mathrm{MSE}\left(\mathbf{\hat{H}}, \mathbf{H}\right) = (N_fN_s)^{-1}\mathbb{E}\left\{\left\Arrowvert\mathbf{\hat{H}} - \mathbf{H}\right\Arrowvert_{F}^{2}\right\}. 
\end{equation}
The bit error ratio (BER) is another metric defined as the ratio of mismatch between the $\mathbf{\hat{s}}$ and $\mathbf{s}$. The hyper-parameters of system configuration are given in Table~\ref{system hyper-parameters}. For training procedure, the training dataset is generated by the design criteria in \cite{luan2025achieving} to realize robust channel estimation, which consists of 125,000 independent channel samples (95\% for training and 5\% for validation). The SNR range is from 5 dB to 25 dB and the $f_{D, max}$ is from 0 Hz to 97 Hz. The loss function for InterpolateNet is the MSE loss, and for the Channelformer, HA02 and proposed method is the Huber loss given in \cite{luan2023channelformer}. To average out Monte Carlo effects, each sample of the simulation curves is tested with 5,000 independent channel realizations. 
\begin{table}[htbp]
\caption{Simulation hyper-parameters}
\begin{center}
\begin{tabular}{|c|c|c|c|}
\hline
\multicolumn{2}{|c|}{Offline-training hyper-parameters} & \multicolumn{2}{c|}{Baseband hyper-parameters} \\ 
\hline
Optimizer& Adam& $N_f$ & 204\\
\hline
Maximum epoch& 100& $N_s$& 14\\
\hline
Initial learning rate (lr) & 0.001& $L_{{\mathrm{CP}}}$& 15\\
\hline
Drop period for lr& 25& $f_{r}$& 5.0GHz\\
\hline
Drop factor for lr& 0.5& $f_{D, max}$ & 97Hz\\
\hline
Minibatch size& 128& $f_{space}$& 15kHz\\
\hline
L2 regularization& $10^{-7}$& SNR range & from 5dB to 25dB \\ 
\hline
\end{tabular}
\label{system hyper-parameters}
\end{center}
\end{table}
\subsection{MSE and BER performance across SNRs}
We evaluate the MSE and BER performance across SNRs from 5dB to 25dB and the maximum Doppler shift is from 0Hz to 97Hz. 
\begin{figure*}
    \centering
    \subfloat[MSE results of the MambaNet \label{MSE}]{%
        \includegraphics[width=0.5\textwidth]{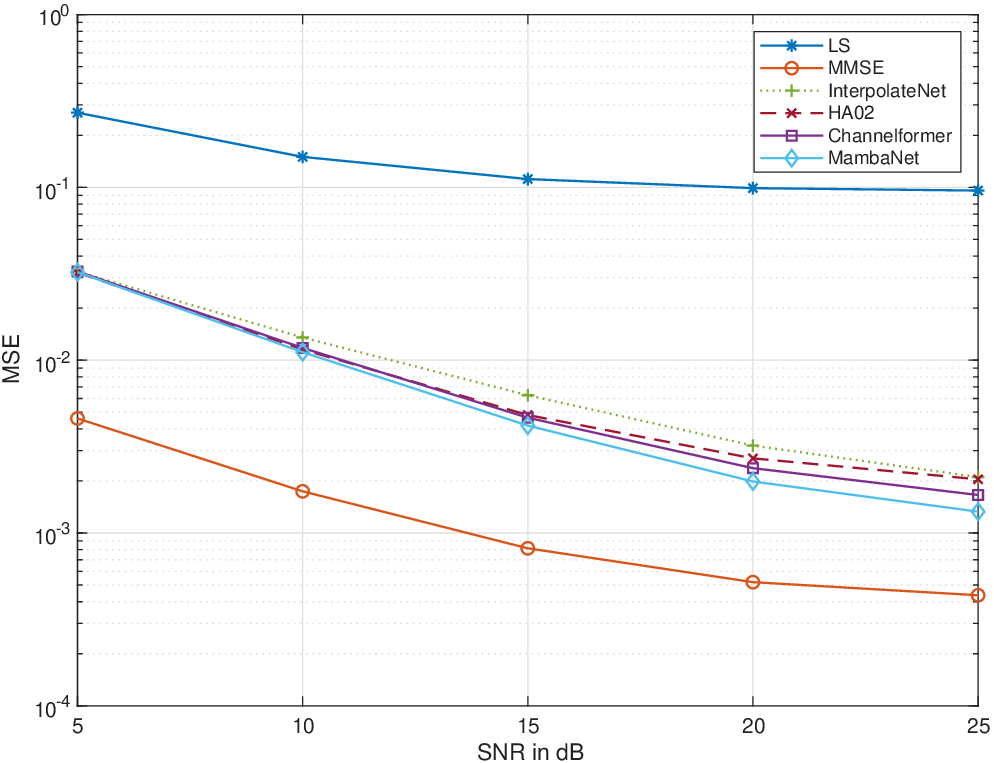}}
    \hfill
    \subfloat[BER results of the MambaNet \label{BER}]{%
        \includegraphics[width=0.5\textwidth]{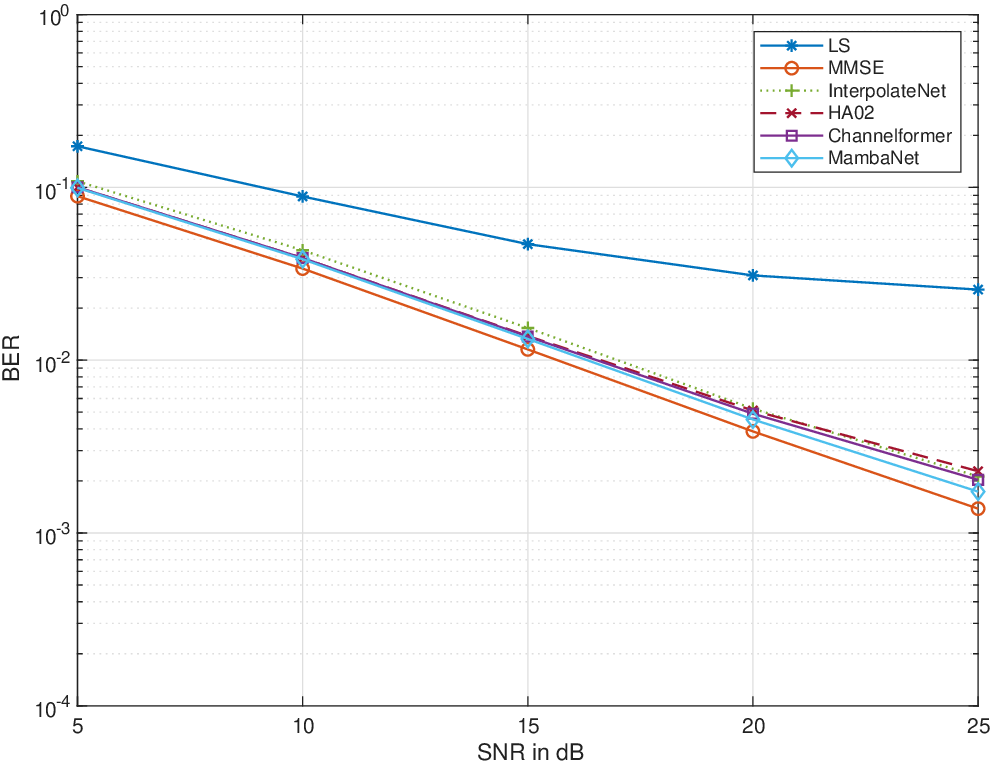}}
  \caption{MSE and BER results for the InterpolateNet, HA02, Channelformer and MambaNet. } 
\end{figure*}

Fig.~\ref{MSE} provides the MSE performance for each method when the design channel matches the test channel. Compared with LS estimate, the MambaNet significantly improves the performance which also outperforms InterpolateNet, HA02 and Channelformer for the whole SNR range. For InterpolateNet, the corresponding MSE varies from 0.0021 to 0.0325 which is worse than other neural network solutions. Compared with HA02 and Channelformer, the MSE of MambaNet decreases by 0.00072 and 0.00033 for 25dB SNR respectively. Moreover, MambaNet outperforms the MMSE method at 25dB SNR, while MMSE requires exact channel information to precisely predict the complete channel matrix. These results demonstrate the effectiveness and practicality of the customized Mamba block, whose bidirectional selective state scan enables long-range propagation of element-wise correlation information, allowing MambaNet to achieve superior estimation performance compared with other solutions. 

Fig.~\ref{BER} compares the BER performance of each method over this extended SNR range, where the MambaNet achieves the best performance. At 5dB SNR point, MambaNet achieves 0.0994 BER while Channelformer has a BER value of 0.1002. The corresponding BER values for InterpolateNet and HA02 are 0.1084 and 0.0999 respectively. The BER values at 25dB SNR are 0.00023 for HA02, 0.00021 for InterpolateNet and 0.0020 for Channelformer respectively. MambaNet achieves a very low BER of 0.0017 while MMSE method only achieves a BER of 0.0014. For the high SNR range, neural network solutions can outperform the MMSE method because this paper uses linear interpolation for the MMSE method to obtain the complete channel prediction. 
\subsection{Validation of generalization property}
This section evaluates the MSE performance across different channels to demonstrate the generalization property of offline-trained MambaNet. The extended pedestrian A (EPA), extended vehicular A (EVA) and extended typical urban (ETU) channels are defined in 3GPP TS 36.101. Flat fading channel has only one path with path gain of 0dB. The two path channel consists of one path with 0ns delay and 0dB gain, and one delayed path with 1000ns and -3.0dB gain. 

\begin{figure}[htbp]
\centerline{\includegraphics[width=0.5\textwidth]{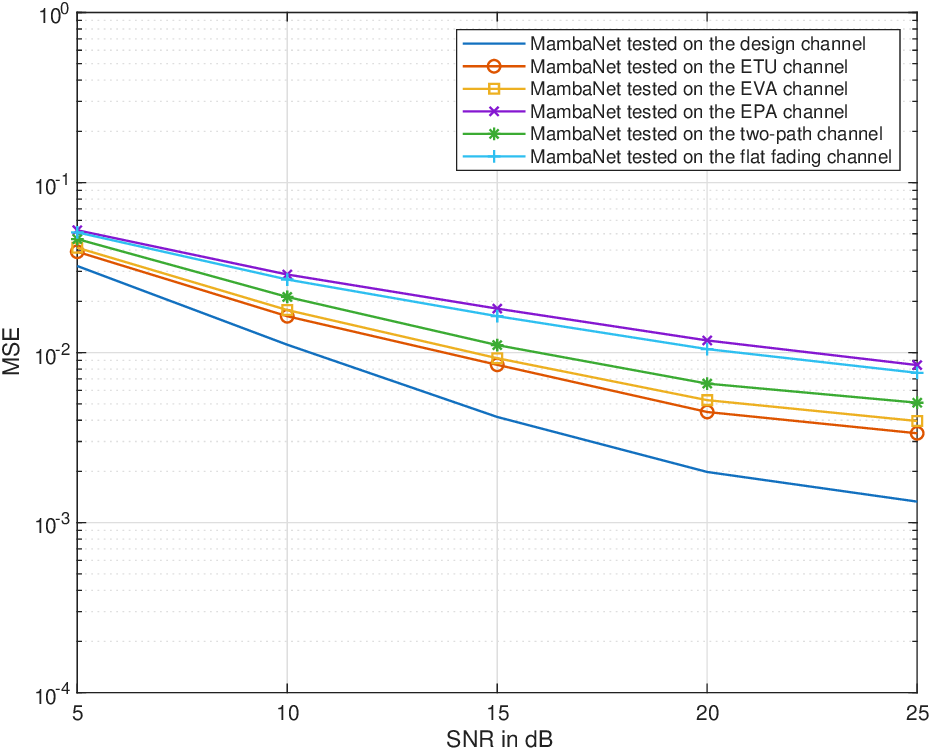}}
\caption{MSE results for offline-trained MambaNet tested on flat fading, two path, EPA, EVA and ETU channels}
\label{Generalization}
\end{figure}

Fig.~\ref{Generalization} shows that offline-trained MambaNet performs robustly across previously unseen channels. However, it does not achieve an almost identical performance as that of the design channel, indicating that the distribution mismatch still leads to moderate performance and generalization degradation. This phenomenon commonly occurs in hybrid neural networks that not only purely contains convolutional neural networks. 
\subsection{Complexity analysis}
Computational complexity of the proposed method and the baseline neural network methods is also compared. With increasing $N_f$, the total number of tunable parameters for MambaNet scales as $\mathcal{O}\left(\left(\frac{N_fN_{pilot}}{L_s}\right)^{2}\right)$. However, for Channelformer and HA02, they exhibit a complexity of $\mathcal{O}\left(\left(N_fN_s\right)^{2}\right)$ which represents a significant increase compared to MambaNet. Both Channelformer and HA02 have severe parameter explosion with increasing $N_f$ to a very large number. However, the number of parameters for MambaNet increases slowly and has a significant reduction. Compared with Channelformer and HA02, MambaNet has a 73\% and 76\% reduction on the number of tunable parameters but requires twice the running time of Channelformer. Although the total number of tunable parameters for interpolateNet is fixed, it shows relatively poor performance in Fig.~\ref{MSE}. Both the number of tunable parameters and the running time normalized by that of LS method for each method are provided in Table~\ref{Complexity analysis}. 
\begin{table}[htbp]
\caption{Complexity analysis}
\begin{center}
\begin{tabular}{c c c}
\hline
 & \textbf{Tunable parameters}& \textbf{Normalized running time} \\
\hline
InterpolateNet& 0.01M & 12.1 \\
\hline
HA02& 0.99M & 15.2 \\
\hline
Channelformer& 0.77M & 22.5 \\
\hline
MambaNet& 0.27M & 24.3 \\
\hline
\end{tabular}
\label{Complexity analysis}
\end{center}
\end{table}
\section{Conclusion}
\label{Conclusion}
In this paper, we propose a hybrid Mamba-attention architecture to improve channel estimation performance for large-scale subcarrier configurations in OFDM systems. The proposed architecture incorporates a customized Mamba module to efficiently capture long distance dependencies across a large number of subcarriers. In particular, a bidirectional state scan is employed to enable the Mamba structure to capture correlation information from both forward and backward directions for inherently non-causal channel features. Simulation results based on 3GPP TS 36.101 channels show that the proposed method achieves improved channel estimation performance with fewer tunable parameters for large-scale subcarriers, while also exhibiting good generalization across previously unseen channels. 
\section*{Acknowledgments}
This research is supported by EPSRC-TITAN projects EP/X04047X/1 and EP/Y037243/1. For the purpose of open access, the author has applied a Creative Commons Attribution (CC BY) licence to any Author Accepted Manuscript version arising from this submission. 
\ifCLASSOPTIONcaptionsoff
  \newpage
\fi

\bibliographystyle{IEEEtran}

\bibliography{Reference}

\end{document}